\documentclass[letterpaper, 10 pt, conference]{ieeeconf}  % Comment this line out if you need a4paper

\IEEEoverridecommandlockouts                              % This command is only needed if 
\usepackage[utf8]{inputenc} % allow utf-8 input
\usepackage[T1]{fontenc}    % use 8-bit T1 fonts
\usepackage[hidelinks]{hyperref}       % hyperlinks
\usepackage{url}            % simple URL typesetting
\usepackage{booktabs}       % professional-quality tables
\usepackage{amsfonts}       % blackboard math symbols
\usepackage{nicefrac}       % compact symbols for 1/2, etc.
\usepackage{microtype}      % microtypography
\usepackage[separate-uncertainty=true,detect-mode=true]{siunitx} % For si units
\usepackage{mathtools}
\usepackage{amssymb}
\usepackage{amsmath}
\usepackage{tabu}
\usepackage{color}
\usepackage{algorithm}
\usepackage{algorithmic}
\usepackage{hyperref}
\usepackage[capitalize]{cleveref}

\usepackage[english]{babel}
\usepackage{graphics}
\usepackage{pgfplots}
\usepackage{pgfplotstable}
\usepackage{tikz}

\usepackage{svg}
\usepackage{subcaption}
\usepackage[style=ieee,backend=bibtex,url=false,doi=false,natbib=true,mincitenames=1,dashed=false,isbn=false]{biblatex}
 \AtEveryBibitem{\clearfield{issn}}
 \AtEveryCitekey{\clearfield{issn}}
 \usepackage{anyfontsize}
\usepackage{amssymb}
\usepackage{amsmath}
\DeclareMathOperator{\E}{\mathbb{E}}
\DeclareMathOperator{\R}{\mathbb{R}}
\DeclareMathOperator*{\argmax}{arg\,max}
\DeclareMathOperator*{\argmin}{arg\,min}

\title{Modeling Human Driving Behavior through \\ Generative Adversarial Imitation Learning}

\author{Raunak Bhattacharyya, Blake Wulfe, Derek J. Phillips, Alex Kuefler, \\ Jeremy Morton, Ransalu Senanayake, Mykel J. Kochenderfer}

\begin{document}
\maketitle
\thispagestyle{empty}
\pagestyle{empty}

\begin{abstract}

An open problem in autonomous vehicle safety validation is building reliable models of human driving behavior in simulation.
    This work presents an approach to learn neural driving policies from real world driving demonstration data.
        We model human driving as a sequential decision making problem that is characterized by non-linearity and stochasticity, and unknown underlying cost functions.
            Imitation learning is an approach for generating intelligent behavior when the cost function is unknown or difficult to specify. 
                Building upon work in inverse reinforcement learning (IRL), Generative Adversarial Imitation Learning (GAIL) aims to provide effective imitation even for problems with large or continuous state and action spaces, such as modeling human driving.
                    This article describes the use of GAIL for learning-based driver modeling. Because driver modeling is inherently a multi-agent problem, where the interaction between agents needs to be modeled, this paper describes a parameter-sharing extension of GAIL called PS-GAIL to tackle multi-agent driver modeling. 
                        In addition, GAIL is domain agnostic, making it difficult to encode specific knowledge relevant to driving in the learning process. 
                            This paper describes Reward Augmented Imitation Learning (RAIL), which modifies the reward signal to provide domain-specific knowledge to the agent. 
                                Finally, human demonstrations are dependent upon latent factors that may not be captured by GAIL. 
                                    This paper describes Burn-InfoGAIL, which allows for disentanglement of latent variability in demonstrations.
                                        Imitation learning experiments are performed using NGSIM, a real-world highway driving dataset. 
                                            Experiments show that these modifications to GAIL can successfully model highway driving behavior, accurately replicating human demonstrations and generating realistic, emergent behavior in the traffic flow arising from the interaction between driving agents.
\end{abstract}

% As a result, learning from human driving demonstrations is a promising approach for generating human-like driving behavior.
\section{Introduction}
\label{Introduction}
Autonomous vehicles have the potential to bring a wide range of benefits in safety, efficiency and access equality. However, comprehensive safety validation is necessary for automated driving systems before their release. Due to their low risk, evaluation time, and cost compared to real-world driving tests, simulations will play a key role in the acceptance of automated driving systems. For such simulations to provide accurate estimates of system performance, they will require representative models of human driver behavior. 
    Further, autonomous vehicles also need robust models of human driving behavior for planning under uncertainty.

Human behavior is responsible for generating the vehicle trajectories that we can observe.
  The properties of these trajectories as observed in real life should be accurately represented by our models.
However, modeling human driving behavior is challenging for multiple reasons. First, human behavior is stochastic, i.e., it is inconsistent across settings and different instants even with all other factors equal. Addressing this inherent stochasticity of human behavior is one of the fundamental challenges in human-robot interaction (HRI). Second, human behavior is multimodal. Even when the broader intent is known, there are multiple actions that a human driver may take. Third, human behavior is influenced by latent factors such as intent and unobserved traits. Fourth, human behavior is governed by high-dimensional states and nonlinearities in the dynamics.

In this paper, we view driving as a sequential decision making problem under uncertainty, and model it as a Markov Decision Process (MDP).
    In order to solve an MDP, a cost function must be specified; however, the cost function is unknown and difficult for driving. 
        Without access to a cost function, imitation learning, also known as learning from demonstration, is a promising approach. Imitation learning uses expert demonstrations to learn a policy that performs similarly to the expert on the unknown cost function.

Traditional approaches to imitation learning have three primary disadvantages. 
    First, they often fail at imitating the expert as a consequence of restricting the class of cost functions. 
        Second, the class of cost functions is often defined as the span of a set of basis functions that must be defined manually (as opposed to learned from observations). 
            Third, they generally involve running reinforcement learning repeatedly, and have a large computational cost as a result.

Generative Adversarial Imitation Learning (GAIL) is a method that tries to address these drawbacks. Using Generative Adversarial Networks (GANs), GAIL removes the restriction that the cost belong to a highly limited class of functions, instead allowing it to be learned using expressive function approximators such as neural networks. 
    Further, using Trust Region Policy Optimization (TRPO), GAIL directly finds policies instead of intermediate value functions.

In this paper, we apply GAIL to real-world driving data in order to learn models of human driving behavior. Our contributions are:
\begin{itemize}
    \item We model human driving behavior using neural network policies that are automatically trained from demonstrations, instead of requiring hand-designed features.
    \item We propose variations to GAIL that are useful specifically for the problem of driver modeling: parameter sharing to enable multi-agent imitation, reward augmentation to provide domain knowledge, and mutual information maximization to uncover individual driving styles.
    \item We demonstrate our driver modeling approach on real world driving demonstration data from the NGSIM driving dataset.
\end{itemize}
\section{Problem Definition}
\label{section:probdef}

\begin{figure*}
\label{fig:imitation_pipeline}
\begin{center}
    \includegraphics[width=2\columnwidth]{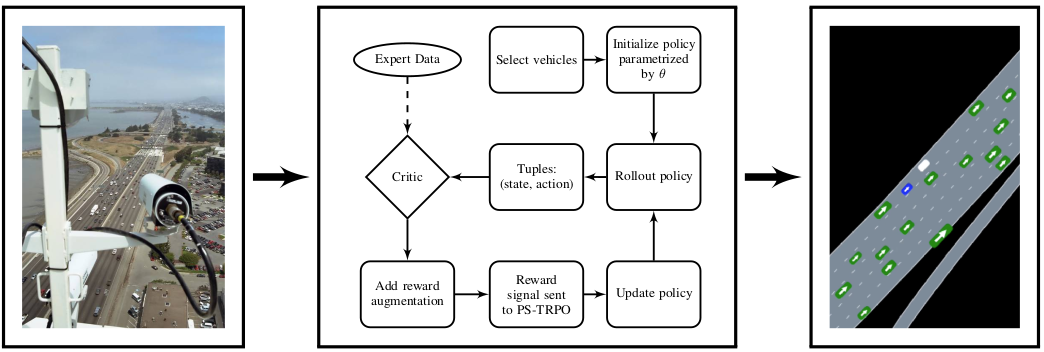}
\end{center}
\caption{The imitation learning pipeline: the demonstration data from the NGSIM dataset (left panel) is fed into our imitation learning module (middle panel) to create driving policies (right panel).}
\end{figure*}

Our goal is to generate realistic multi-agent behaviors automatically for traffic simulation.
    We define the multi-agent human driving behavior modeling problem as learning automatic driving policies from human demonstrations that generate realistic car-following and lane-changing behavior for all the simulated vehicles on the road segment of interest.
      Human behavior (e.g., their propensity to change lanes as a result of a car in front of them slowing down) is responsible for generating the vehicle trajectories that we can observe.
        We model each traffic scenario as a sequential process where traffic actors interact and plan their behaviors at each timestep.

The state space represents the driving scene, the actions are driving actions, and the transition model is governed by the vehicle dynamics and the actions taken by surrounding vehicles. 
However, the cost function of the MDP is unspecified because it is often difficult for humans to articulate, let alone mathematically formulate, the cost function that they are following while driving.
Given a class of policies $\pi_\theta$ parameterized by $\theta$, we seek to find the policy that best recreates human driving behavior.
The goal is to infer this policy from a dataset consisting of a sequence of state-action tuples $(s_t, a_t)$.

We use the public Next-Generation Simulation (NGSIM) dataset for US Highway 101~\citep{colyar2007us}.
NGSIM provides \num{45} minutes of driving at \SI{10}{Hz}.
The dataset covers an area in Los Angeles approximately \SI{640}{m} in length with five mainline lanes and a sixth auxiliary lane for highway entrance and exit.

Traffic density in the dataset transitions from uncongested to full congestion and exhibits a high degree of vehicle interaction as vehicles merge on and off the highway and must navigate in congested flow.
The diversity of driving conditions and the forced interaction of traffic participants makes these sources particularly useful for behavioral studies.
The trajectories were smoothed using an extended Kalman filter on a bicycle model and projected to lanes using centerlines extracted from the NGSIM roadway geometry files.
Cars, trucks, buses, and motorcycles are in the dataset, but only car trajectories were used for model training.

This dataset is split into three consecutive 15 min sections of driving data that represent different vehicle densities and traffic conditions.
We use the first section as the training dataset, from which we learn our policies.
The remaining two sections are used for testing and evaluating the quality of the resulting policies.
This allows us to assess the generalization capability of the learned driving policies.
\section{Related Work}
Existing approaches to multi-agent human driving behavior modeling include open-loop independent trajectory prediction, game theoretic prediction, and closed-loop forward simulation.

Under the open-loop independent trajectory prediction paradigm, models predict the trajectory independently for each agent in the scene.
  Some simple models in this paradigm are based on various combinations of constant velocity, acceleration, yaw rate or steering angle.
    More advanced models rely on deep neural networks~\citep{altche2017lstm,deo2018would,kim2017probabilistic,krajewski2019beziervae,yoon2016multilayer}, Gaussian processes~\citep{armand2013modelling} and Gaussian mixture models~\citep{wiest2012probabilistic}.
      Because these models ignore interaction between agents, their predictive power diminishes with increasing prediction horizon.

The trajectory prediction problem has also been studied in the computer vision community. 
  To characterize the behavior of multiple actors jointly, auto-regressive generation with social mechanisms have been used~\citep{alahi2016social,rhinehart2019precog,suo2021trafficsim,tang2019multiple}. 
    However, these cannot directly be used for simulation over long time horizons because they are sensitive to distributional shift and may not recover from compounding error.

Under the game-theoretic paradigm, the predicted motion of some agents is explicitly conditioned on the predicted motion of other agents in the scene~\citep{pruekprasert2019decision,okuda2017design,isele2019interactive,fisac2019hierarchical,gonzalez2019human}. 
  Thus, agents are modeled as looking ahead to consider the ramifications of their actions. 
    This notion of looking ahead makes game theoretic models more deeply interaction-aware than forward simulation models. 
      However, game theoretic models suffer from tractability issues making them unsuitable for building traffic simulators.

In the closed-loop forward simulation paradigm, the model computes a control action for each agent at each time step based on the observations received up to and including that time step, and it then propagates the entire scene forward in time~\citep{abbeel2004apprenticeship,gipps1981behavioural,kuefler2017imitating,levine2012continuous,morton2016analysis}.
  Since forward simulation uses closed-loop policies, models are nominally interaction-aware. 
    The control action can depend on the actions of other agents through the observation received by the ego vehicle.
      Microscopic traffic simulators~\citep{treiber2000congested,kesting2007general} employ heuristic models to simulate traffic flow. 
        While these may capture high-level traffic characteristics by directly encoding traffic rules, they typically do not not capture realistic interactions, motivating the use of imitation learning.
\section{Imitation Learning}
\label{section:background}
An infinite horizon, discounted MDP is defined by the tuple $(\mathcal{S}, \mathcal{A}, P, c, \rho_0, \gamma)$, where $\mathcal{S}$ is the state space, $\mathcal{A}$ is the action space, $P:\mathcal{S} \times \mathcal{A} \times \mathcal{S} \to \R$ is the transition probability distribution, $c: \mathcal{S} \times \mathcal{A} \to \R$ is the cost function, $\rho_0: \mathcal{S} \to \R$ is the distribution of the initial state $s_0$, and $\gamma \in (0,1)$ is the discount factor. 

% The transition model $P(s' \mid s,a)$ governs the successor state when an action $a$ is obtained by a policy $\pi(a \mid s)$. The discount factor governs how much future rewards are valued relative to immediate rewards.

A stochastic policy, $\pi: \mathcal{S} \times \mathcal{A} \to [0,1]$, defines the probability of taking each action from each state. The set $\Pi$ contains all stationary stochastic policies that take actions in $\mathcal{A}$ given states in $\mathcal{S}$. We use $\pi_E$ to refer to the expert policy. In practice, $\pi_E$ will only be provided as a set of trajectory samples obtained by executing $\pi_E$ in the environment. 
% The sum of discounted costs, or return, from state $s_{t}$ is defined as $g_{t} = \sum_{k=0}^{\infty}\gamma^{k}c_{t+k+1}$, where $t$ is a time index, and $c_{t}$ is the corresponding cost. The objective in an MDP is to find a policy that minimizes the expected return, or value, of each state $V_{\pi}(s) = \E_{\pi}[g_{t}\mid s_{t}=s]$.

The expectation with respect to a policy $\pi \in \Pi$ is used to denote an expectation with respect to the trajectory it generates: $\E_{\pi}[c(s,a)] = \E\big[\sum_{t=0}^{\infty}\gamma^t c(s_t,a_t) \big]$ where $s_0 \sim p_0$, $a_t \sim \pi(\cdot \mid s_t)$, and $s_{t+1}\sim P(\cdot \mid s_t,a_t)$ for $t \geq 0$. 
The $\gamma$-discounted causal entropy of the policy $\pi$ is $H(\pi) = \E_{\pi}[-\log \pi(a \mid s)]$. 

The state-occupancy distribution: 
\begin{equation}
\rho_{\pi}(s) = (1-\gamma)\sum_{t=0}^{\infty} \gamma^{t} p(s_{t}=s \mid \pi) \text{,}
\end{equation}
gives the average discounted probability of the agent being in state $s$. The state-action occupancy distribution of a policy $\pi$ is then defined as $\rho_{\pi}(s,a)=\pi(a\mid s)\rho_{\pi}(s)$. This can be interpreted as the distribution of state and actions that an agent encounters when following policy $\pi$ starting from state $s_0 \sim \rho_0$. The state-action occupancy distribution allows us to write the expected trajectory cost of a policy as
\begin{equation}
    \E_{\pi}[c(s,a)] = \E\big[\sum_{s,a} \rho_{\pi}(s,a)c(s,a) \big]\text{,}
\end{equation}
for any cost function $c$.

% \section{Imitation Learning}
% \label{section:il}
The goal of imitation learning (IL) is to learn a policy $\pi$ that imitates an expert policy $\pi_E$ given demonstrations from that expert \citep{schaal1999imitation,ross2011reduction}. A demonstration is defined as a sequence of state-action pairs that result from a policy interacting with the environment: $\tau = \{s_{1},a_{1},s_{2},a_{2},\ldots \}$. 

Behavioral cloning learns a policy by minimizing some loss function $\ell$ over the set of demonstrations with respect to the policy~\citep{ross2011reduction}:
\begin{equation}
\label{eq:sup_obj}
\pi_{sup} = \underset{\pi}{\text{argmin}} \E_{s \sim \rho_{\pi_{E}}}[\ell(\pi,s)] \text{,}
\end{equation}
where $\ell$ is typically the cross-entropy loss when using discrete actions and the negative log likelihood of a multivariate Gaussian distribution when using continuous actions.

During training, behavioral cloning samples states from the state-occupancy distribution of the expert, $\rho_{\pi_{E}}$. However, when interacting with the environment, the policy samples states from the state-occupancy distribution of the learned policy, $\rho_{\pi_{sup}}$. 
    This change in distribution between training and test time is called covariate shift \citep{shimodaira2000improving}, and results in the agent making increasingly large errors. 

Allowing the agent to interact with the environment at training time addresses the underlying cause of covariate shift, but this interaction requires an explicit or implicit reward function since the agent may encounter states not contained in the training data. There are various approaches to addressing this problem, which we detail next.

\subsection{Apprenticeship Learning}
\label{section:al}
The goal of apprenticeship learning \citep{abbeel2004apprenticeship} is to find a policy that performs no worse than the expert under the true cost function:

\begin{equation}
\label{eqn:true_c}
    \E_{\pi}[c^\mathrm{true}(s,a)] \leq \E_{\pi_E}[c^\mathrm{true}(s,a)] \text{.}
\end{equation}

The problem is that the true cost function $c^\mathrm{true}$ is unknown. Hence, the desired goal is recast as:
\begin{equation}
\label{eqn:all_c}
    \E_{\pi}[c(s,a)] \leq \E_{\pi_E}[c(s,a)], \; \forall c \in \mathcal{C}
\end{equation}
where $\mathcal{C}$ is a restricted class of cost functions. Under the assumption that $c^\mathrm{true} \in \mathcal{C}$, if the goal in \cref{eqn:all_c} is met, the policy also satisfies the goal established in \cref{eqn:true_c}.

If we can satisfy \cref{eqn:all_c} for the worst possible cost function, i.e., find a policy that performs no worse than the expert on the worst possible cost function in $\mathcal{C}$, we can guarantee that it will perform no worse than the expert on the (unknown) true cost function. Thus, for a given policy $\pi$ that is yet to be determined, we are interested in finding the worst possible cost function. This is made possible by posing the following optimization problem:
\begin{equation}
\label{eqn:worst_case_cost}
    c^\mathrm{worst}(s,a) = \max_{c \in \mathcal{C}} \E_{\pi}[c(s,a)] - \E_{\pi_E}[c(s,a)] \text{.}
\end{equation}

Once the worst-case cost function $c^\mathrm{worst}$ is known, finding a policy can be posed as the following optimization problem:
\begin{equation}
\label{eqn:policy_find}
    \pi = \argmin_{\pi \in \Pi} \E_{\pi}[c^\mathrm{worst}(s,a)] \text{.}
\end{equation}

The policy found from \cref{eqn:policy_find} is guaranteed to perform no worse than the expert with respect to the worst-case cost function, and hence guaranteed to perform no worse than the expert on the true cost function $c^\mathrm{true}$ if $c^\mathrm{true} \in \mathcal{C}$.

We can add the expert incurred cost into the objective function without changing the resulting optimum, as follows:
\begin{equation}
    \pi = \argmin_{\pi \in \Pi} \E_{\pi}[c^\mathrm{worst}(s,a)]- \E_{\pi_E}[c^\mathrm{worst}(s,a)] \text{.}
\end{equation}

Since the worst-case cost function is found by solving a maximization problem in eq. (\ref{eqn:worst_case_cost}), the overall objective function can be rewritten as:
\begin{equation}
\label{eqn:apprenticeship_learning}
    \pi = \argmin_{\pi \in \Pi} \max_{c\in\mathcal{C}}\E_{\pi}[c(s,a)]- \E_{\pi_E}[c(s,a)] \text{.}
\end{equation}

\Cref{eqn:apprenticeship_learning} establishes a general framework for defining apprenticeship learning algorithms. To use this framework, we need a cost function class $\mathcal{C}$, and an optimization algorithm.

The unknown, true cost function is typically assumed to be a linear combination of known functions that are called basis cost functions. Classic apprenticeship learning algorithms~\citep{abbeel2004apprenticeship,syed2008game} restrict $\mathcal{C}$ to convex sets given by linear combinations of basis cost functions. 
    However, if the true cost function does not belong to the cost function class, we cannot guarantee that our agent will perform no worse than the expert.
\section{Generative Adversarial Imitation Learning}
\label{section:gail}
Generative Adversarial Imitation Learning is derived from an alternative approach to imitation learning called Maximum Causal Entropy IRL (MaxEntIRL)~\citep{bloem2014infinite,ziebart2008maximum}. While apprenticeship learning attempts to find a policy that performs at least as well as the expert across cost functions, MaxEntIRL seeks a cost function for which the expert is uniquely optimal. This latter objective turns out to be equivalent, under certain assumptions, to finding a policy with an occupancy distribution matching that of the expert. This section describes the derivation of this connection, the resulting imitation learning algorithm, and its connection with Generative Adversarial Networks.

\subsection{Derivation of GAIL}
GAIL is derived from a cost-regularized MaxEntIRL objective \citep{ziebart2010modeling}: 

\begin{equation}
\begin{aligned}
\label{eqn:irl_primitive}
    \text{IRL}_{\psi}(\pi_E) = \argmax_{c \in \mathcal{C}} -\psi(c) + 
    \left(\min_{\pi \in \Pi} -\mathcal{H}(\pi) +  \E_{\pi}[c(s,a)]\right) \\
    - \E_{\pi_E}[c(s,a)]\text{.}
\end{aligned}
\end{equation}
where $\mathcal{H}(\pi) \equiv \E_{\pi}[-\log\pi(a \mid s)]$ is the discounted causal entropy of the policy taken with respect to the state-action distribution of the policy, and $\psi: \mathcal{C}\to \R^*$ is a function assigning a value in the extended reals to each cost function $c$. The regularization function, $\psi$, plays an important role in the derivation of GAIL and in its connection with the apprenticeship learning methods of \cref{section:al}. Specifically, \citet{ho2016generative} characterize the result of running reinforcement learning on a cost output from MaxEntIRL:

\begin{equation}
\label{eqn:prop_32}
    \text{RL}\circ\text{IRL}_{\psi}(\pi_E) = \argmin_{\pi \in \Pi} 
    -\mathcal{H}(\pi) + 
    \psi^*(\rho_{\pi}-\rho_{\pi_E})\text{.}
\end{equation}
Here, $\psi^*(\rho_{\pi}-\rho_{\pi_E}) \equiv \sup_{c \in \mathcal{C}} (\rho_{\pi}-\rho_{\pi_E})^T c - \psi(c)$ denotes the convex conjugate of $\psi$, which attempts to find a cost function that places high cost on state-action pairs more frequently visited by $\pi$ than by $\pi_{E}$. As a result, minimizing with respect to $\pi$ attempts to match the occupancy distributions of the two policies.

\citet{ho2016generative} show that different cost function regularizers result in different imitation learning algorithms. For example, they show (under assumptions) that when $\psi$ is constant across cost functions, this results in exact occupancy distribution matching. This is accomplished by showing that MaxEntIRL is dual to the following optimization problem:

\begin{equation}
\label{eqn:primal_problem}
\begin{aligned}
    & \min_{\rho \in \mathcal{D}} -\bar{H}(\rho) \; 
    \text{subject to} \; \rho(s,a) &= \rho_E(s,a), \; \forall s \in \mathcal{S}, a \in \mathcal{A}\\
\end{aligned}
\end{equation}
where $\bar{H}$ denotes the entropy of the occupancy distribution. Solving this optimization problem is intractable for large MDPs because it involves satisfying a constraint for each point in $\mathcal{S} \times \mathcal{A}$, many of which will require $\rho_{\pi}$ to be zero due to the limited size of the dataset of expert demonstrations.

An alternative setting of the cost function regularizer results in the apprenticeship learning algorithms from \cref{section:al}. Let $\psi(c) = \delta_{\mathcal{C}}(c)$ where $\delta_{\mathcal{C}}(c) = 0\; \textrm{if}\; c \in \mathcal{C}$ and $\infty$ otherwise, for a restricted class of cost functions $\mathcal{C}$. This results in \cref{eqn:prop_32} reducing to (entropy-regularized) apprenticeship learning as follows:
\begin{equation}
\label{eqn:reduce2al}
\begin{aligned}
    \pi^\mathrm{app} &= \argmin_{\pi \in \Pi}
    -\mathcal{H}(\pi) +
    \psi^*(\rho_{\pi}-\rho_{\pi_E}) \\
    &= \argmin_{\pi \in \Pi}
    -\mathcal{H}(\pi) +
    \delta_{\mathcal{C}}^*(\rho_{\pi}-\rho_{\pi_E}) \\
    &= \argmin_{\pi \in \Pi} 
    -\mathcal{H}(\pi) +
    \max_{c \in \mathcal{C}} -\delta_{\mathcal{C}}(c) + \\
    & \sum_{s,a}c(s,a)\big(\rho_{\pi}(s,a) - \rho_{\pi_E}(s,a) \big) \\
    & = \argmin_{\pi \in \Pi} 
    -\mathcal{H}(\pi) +
    \max_{c \in \mathcal{C}}  
    \E_{\pi}[c(s,a)] - \\
    & \E_{\pi_E}[c(s,a)]\\
\end{aligned}
\end{equation}

This regularizer restricts the cost function to $\mathcal{C}$, which is traditionally taken to be a small subspace spanned by finitely many basis cost functions.
From \cref{eqn:prop_32} we see that the information contained in the expert policy (or demonstrations sampled using that policy) must be encoded in the cost function. When the ``true'' cost function is not in this space, information about the expert policy can be lost, which partially explains why traditional apprenticeship learning algorithms can fail to imitate the expert well.

Given that the desire is for an imitation learning algorithm that can: 1) scale to large state action spaces to work for practical problems, and 2) can allow for imitation without restricting cost functions to lie in a small subspace of finitely many linear basis cost functions, GAIL proposes a new cost function regularizer $\psi_{GA}$. This regularizer allows scaling to large state action spaces and removes the requirement to specify basis cost functions. While existing apprenticeship learning formalisms used the cost function as the descriptor of desirable behavior, GAIL relies instead on the divergence between the demonstration occupancy distribution and the learning agent's occupancy distribution. The subsequent discussion will derive the form of $\psi_{GA}$ and establish its connection to GANs.

\subsection{Connection to Generative Adversarial Networks}
\Cref{eqn:prop_32} establishes an optimization problem formulation for the imitation learning problem via the convex conjugate of the cost function regularizer.
A cost function regularizer $\psi(c)$ needs to be instantiated to reach an imitation learning algorithm. Consider the binary classification problem of classifying state-action pairs $(s,a)$ that have been drawn from the expert occupancy distribution $\rho_{\pi_E}$ or the learning agent's occupancy distribution $\rho_\pi$. For this binary classification task, assuming a loss function $\phi$ to score the training examples, the minimum expected risk is defined as:
\begin{equation}
\label{eqn:min_expected_risk}
R_{\phi}(\pi,\pi_E) = \sum_{s,a}\min_{\gamma \in \R} \rho_{\pi}(s,a)\phi(\gamma) + \rho_{\pi_E}(s,a)\phi(-\gamma) \text{.}
\end{equation}
Proposition A.1 of \citet{ho2016generative} shows that this minimum expected risk is connected to the convex conjugate of the cost function regularizer as
\begin{equation}
\label{eqn:psi_minrisk}
    \psi_{\phi}^*(\rho_{\pi}-\rho_{\pi_E}) = -R_{\phi}(\rho_{\pi},\rho_{\pi_E}) \text{.}
\end{equation}
This allows us to build a bridge from binary classification to imitation learning. The logistic loss function $\phi(x) = \log(1+e^{-x})$ is chosen. In this case, using the logistic loss, the connection is (as shown by corollary A.1.1 from \citep{ho2016generative}):
\begin{equation}
\label{eqn:minrisk_gan}
\begin{aligned}
    \psi_{GA}^*(\rho_{\pi}-\rho_{\pi_E}) &= -R_{\phi}(\rho_{\pi},\rho_{\pi_E})\\
    &= \sum_{s,a}\max_{\gamma \in \R} \rho_{\pi}(s,a)\big(-\phi(\gamma)\big) + \\
    & \rho_{\pi_E}(s,a)\big(-\phi(-\gamma)\big)\\
    &= \sum_{s,a}\max_{\gamma \in \R} \rho_{\pi}(s,a) \log\bigg(\frac{1}{1+e^{-\gamma}} \bigg) + \\ & \rho_{\pi_E}(s,a)\log\bigg(\frac{1}{1+e^{\gamma}}\bigg)\\
    &= \sum_{s,a}\max_{\gamma \in \R} \rho_{\pi}(s,a) \log\big(\sigma(\gamma) \big) + \\
    & \rho_{\pi_E}(s,a)\log\big(1-\sigma(\gamma)\big)\\
    &= \sum_{s,a}\max_{d \in (0,1)} \rho_{\pi}(s,a) \log d + \\ & \rho_{\pi_E}(s,a)\log(1-d)\\
    &= \max_{D} \sum_{s,a}\rho_{\pi}(s,a)\log(D(s,a))+\\
    & \rho_{\pi_E}(s,a)\log(1-D(s,a))\text{.}
\end{aligned}
\end{equation}

If we use the result from \cref{eqn:minrisk_gan} to substitute the expression for the convex conjugate of the cost regularizer into our central optimization objective \cref{eqn:prop_32}, we obtain the GAIL objective function:
\begin{equation}
\label{eqn:gail}
    \min_{\pi} \max_{D} \sum_{s,a}\rho_{\pi}(s,a)\log(D(s,a))+\rho_{\pi_E}(s,a)\log(1-D(s,a))\text{.}
\end{equation}

This optimization objective established in \cref{eqn:gail} provides a connection to GANs \citep{goodfellow2014generative}. In GANs, the goal is to model the distribution $p_\mathrm{data}(x)$. The generative modeling objective is formulated as
\begin{equation}
    \min_{G}\max_{D} \E_{x \sim p_\mathrm{data}(x)}[\log D(x)]+\E_{z \sim p_z(z)}[\log (1-D(G(z)))]\text{.}
\end{equation}

Here, $G$ is the generator that maps input noise variables $z$ to the data space as $G(z)$ and $D$ is the discriminator which outputs a single scalar $D(x)$ that represents the probability that $x$ came from the data rather than $p_g$, a binary classification task. 
    This objective is solved using simultaneous gradient descent wherein the parameters of $G$ and $D$ are updated by sampling two sets of data, one from the training samples and the other from the noise prior.

Unlike GANs, GAIL considers the environment as a black box, and thus the objective is not differentiable with respect to the parameters of the policy. Therefore, simultaneous gradient descent is not suitable. Instead, optimization over the GAIL objective is performed by alternating between a gradient step to increase \cref{eqn:gail} with respect to the discriminator parameters $D$, and a Trust Region Policy Optimization (TRPO) step \citep{schulman2015trust} to decrease \cref{eqn:gail} with respect to the parameters $\theta$ of the policy $\pi_{\theta}$.

% GAIL can also be derived more directly from a $f$-divergence minimization perspective \citep{ghasemipour2020divergence}, which is less general than cost-regularized MaxEntIRL. The $f$-divergence framework does not allow for minimizing certain distances between occupancy distributions, for example the Wasserstein distance, which has been shown to result in more reliable training of GANs \citep{arjovsky2017wasserstein}. However, the Wasserstein distance can be used within the cost-regularized MaxEntIRL framework \citep{xiao2019wasserstein}, and we use this version of GAIL in the multi-agent setting.
\subsection{Information Maximizing GAIL}
Demonstration trajectories are typically collected from human experts. However, these trajectories can show significant variability due to internal latent factors of variation among different individuals. For example, aggressive drivers will demonstrate significantly different driving trajectories as compared to passive drivers, even for the same road geometry and traffic scenario. To uncover these latent factors of variation, and learn policies that produce trajectories corresponding to these latent factors, Information Maximizing GAIL (InfoGAIL) was proposed~\citep{li2017infogail}.

InfoGAIL assumes that the expert policy is a mixture of experts $\pi_E = \{\pi_E^0,\pi_E^1,...\}$, and defines the generative process of the expert trajectory $\tau_E$ as $s_0 \sim \rho_0, z \sim p(z), \pi \sim p(\pi \mid z), a_t \sim \pi(a_t \mid s_t), s_{t+1} \sim P(s_{t+1} \mid a_t,s_t)$, where $z$ is a discrete latent variable that selects a specific policy $\pi$ from the mixture of expert policies $p(\pi \mid z)$ (which is unknown and needs to be learned), and $p(z)$ is the known prior distribution of $z$.

In the GAIL formulation, there is no incentive given to separating and disentangling variations observed in the data. The latent variable $z$ is introduced for this purpose. To ensure that the learned policy utilizes $z$ as much as possible, InfoGAIL tries to enforce high mutual information between the latent variable $z$ and the state-action pairs in the generated trajectory given by
\begin{equation}
\label{eqn:mutual_info}
    I(z;\tau) = H(z) - H(z \mid \tau)\text{.}
\end{equation}
Intuitively, the mutual information captures the amount of information obtained from knowledge of the trajectory $\tau$ about the latent variable $z$.

However, capturing the mutual information in \cref{eqn:mutual_info} relies on knowledge of the probability distribution $P(z \mid \tau)$, which is difficult to access. Therefore, a variational lower bound $L_I(\pi,Q)$, of the mutual information $I(z;\tau)$ is introduced \citep{chen2016infogan}, where $Q(z \mid \tau)$ is an approximation of the true posterior $P(z \mid \tau)$. This lower bound is given by 
\begin{equation}
\begin{aligned}
\label{eqn:lower_bound}
    L_I(\pi,Q) &= \E_{z \sim p(z),a \sim \pi(\cdot \mid s,z)}[\log Q(z \mid \tau)] + H(z) \\
               & \leq I(z;\tau)\text{.}
\end{aligned}
\end{equation}

Now, the GAIL policy learning objective function under this mutual information regularization is modified to
\begin{multline}
\label{eqn:infogail}
    \min_{\pi,Q} \max_{D} \E_{\pi}[\log D(s,a)] + \E_{\pi_E}[\log (1-D(s,a))] - \\
    \lambda L_I(\pi,Q)\text{,}
\end{multline}
where $\lambda$ is the hyperparameter for the information maximization regularization term. In \cref{eqn:infogail}, the latent code capturing the variability in demonstration is introduced via $Q(z \mid \tau)$, the approximation to the posterior distribution $P(z \mid \tau)$.

However, if the policy is initialized from a state sampled at the end of a demonstrator's trajectory (as is the case when initializing the ego vehicle from a human playback), the driving policy's actions should be consistent with the driver's past behavior. InfoGAIL relies on sampling a random latent code at the beginning of a trial, which cannot ensure the requirement of consistency with the true driving style. This shortcoming limits the applicability of InfoGAIL to modeling real driving situations, where ego vehicles are sampled from playbacks of recorded human data.

To address this issue of inconsistency with real driving behavior, Burn-InfoGAIL~\citep{kuefler2018burn} was introduced, where a policy must take over where an expert demonstration trajectory ends. This is referred to as a burn-in demonstration, upon which a learned inference model must be conditioned to draw latent codes that characterize driving style.
\subsection{Extension to Multiple Agents}
\label{section:multi-agent}
Our goal is to simulate the behavior of not just a single vehicle, but entire traffic scenes to be able to recreate driving behavior arising out of interaction between agents. 
The task of simultaneously controlling multiple vehicles operating on a single roadway can be viewed as a multiagent control problem.
However, when evaluated in a multi-agent setting, the policies learned through single-agent imitation learning fail to exhibit realistic behavior, rendering them inadequate for use in simulation. 
During test time, the policy learned from a single agent's experience observes nearby vehicles acting differently than during training, and again makes small errors that compound over time. 

This motivated the development of Parameter-Sharing GAIL (PS-GAIL) \citep{bhattacharyya2018multi}, which enables scaling of the imitation learning approach to multiple agents.
In line with recent work in multi-agent imitation learning \citep{song2018multi,yu2019multi,gruver2020multi}, we formulate multi-agent driving as a Markov game \citep{littman1994markov} consisting of $M$ agents and an unknown reward function. We make three simplifying assumptions:

\begin{enumerate}
\item Homogeneous agents: agents have the same observation and action spaces:
\begin{align*}
 \mathcal{O}_{i} = \mathcal{O}_{j} \text{ and } \mathcal{A}_{i} = \mathcal{A}_{j} \text{ } \forall \text{ agents } i,j \text.
\end{align*}

\item Identical cost function: the (unknown) cost function is the same for all agents:
\begin{align*}
\mathcal{R}_{i} = \mathcal{R}_{j} \text{ } \forall \text{ agents } i,j \text.
\end{align*}

\item Cooperative agents: agents have common behavior such as actions, domain knowledge, and goals (an example is avoiding collision). In particular, drivers are not adversarial (as in the case of racing).

% \item Independent rewards: the reward function is not shared; it depends only on the action of each agent and the state, and not on the actions of other agents or the next state. In particular, agents are not cooperative:
% \begin{align*}
% \mathcal{R}_{i}(s, a_{1}, \ldots, a_{i}, \ldots, a_{k}) = \mathcal{R}_{i}(s, a_{i})\text.
% \end{align*}

\end{enumerate}

These assumptions are idealizations and do not hold for real-world driving scenes. For example, different vehicles may permit different accelerations, a driver may only want to change lanes if other drivers are not doing so, and individuals may value different driving qualities such as smoothness or proximity to other vehicles differently. Nevertheless, these assumptions often do apply approximately, and, as we later show, allow for learning of realistic driving policies.

A naive approach to learning human driver policies would be to train a policy in an environment where it controls a single vehicle on the roadway and all remaining vehicles follow a predetermined trajectory.
Unfortunately, this approach is often incapable of producing policies that can reliably control many vehicles on the same roadway.
By introducing such a controller to other vehicles after training, we reintroduce covariate shift. As a result, small errors in the behavior of a single vehicle can destabilize neighboring vehicles, ultimately leading to the failure of many agents in the scene. 

Existing training schemes for reinforcement learning that are applicable to multi-agent domains are centralized learning~\citep{foerster2016learning,sukhbaatar2016learning} and concurrent learning~\citep{diallo2017learning,palmer2018lenient}. Centralized training learns a single policy that controls every agent by mapping a joint observation to a joint action over all the agents. However, this quickly runs into scalability issues with increasing number of agents due to exponential growth in the observation and action spaces. In the concurrent training approach, every agent learns its own policy. This addresses the growth of the state and action spaces, but still scales poorly because the number of parameters to be learned grows with the number of agents.

Centralized learning and decentralized execution is an approach that finds the middle ground between centralized and concurrent approaches to learning in multi-agent domains. Parameter sharing is an example of this approach that has shown recent promise due to more centralization during the learning process~\citep{terry2020parameter}.
Further, direct policy search based methods have shown to perform better than value based methods in deep reinforcement learning problems involving continuous action spaces~\citep{nguyen2020deep}, such as driving.

Combining the benefits of parameter sharing with direct policy search, \citet{gupta2017cooperative} introduced an algorithm called Parameter Sharing Trust Region Policy Optimization (PS-TRPO), which is a policy gradient approach that combines parameter sharing and TRPO.
PS-TRPO was shown to produce decentralized parameter-sharing neural network policies that exhibit emergent cooperative behavior without explicit communication between agents. PS-TRPO is highly sample-efficient because it reduces the number of parameters by a factor of $M$, and shares experience across all agents. Furthermore, it mitigates issues resulting from non-stationary learning dynamics by collecting a new batch of data for each round of policy optimization. Notably, PS-TRPO still allows agents to exhibit different behavior because each agent receives unique observations.

For a policy $\pi_{\theta}$ with parameters $\theta$, PS-TRPO performs an update to the policy parameters by approximately solving the constrained optimization problem:
\begin{equation}
\label{eq:trpoobj}
\begin{aligned}
& \underset{\theta}{\text{maximize}}
& & \E_{o,a \sim \pi_{\theta_k}} \left[{\frac{\pi_{\theta}(a \mid o)}{\pi_{\theta_k}(a \mid o)} A_{\theta_k}(o, a)} \right] \\
& \text{subject to}
& & \E_{o}\left[{D_{KL}(\pi_{\theta_k}(\cdot \mid o) \Vert \pi_{\theta}(\cdot \mid o))}\right] \leq \Delta_{KL} \text{,}
\end{aligned}
\end{equation}
where $\pi_{\theta_k}$ is a rollout-sampling policy, and $A_{\theta_k}(o, a)$ is an advantage function quantifying how much the value of an action $a$ taken in response to an observation $o$ differs from the baseline value estimated for $o$. $D_{KL}$ is the KL-divergence between the two policy distributions, and $\Delta_{KL}$ is a step size parameter that controls the maximum change in policy per optimization step.

Our approach, PS-GAIL, combines GAIL with PS-TRPO to generate policies capable of controlling multiple vehicles, enabling more stable simulation of entire road scenes.
The approach is described in ~\cref{algo:psgail}. We begin by initializing the shared policy parameters and select a step size parameter.
At each iteration, the shared policy is used by each agent to generate trajectories. Rewards are then assigned to each state-action pair in these trajectories by the critic. 
Subsequently, observed trajectories are used to perform a TRPO~\citep{schulman2015trust} update for the policy, and an Adam~\citep{kingma2014adam} update for the critic.
PS-GAIL can be viewed as a special case of the algorithms presented by Song et al.~\citep{song2018multi} since all agents share the same policy and receive rewards from the same critic.

% Algorithm: Parameter sharing GAIL
\begin{algorithm}
  \caption{PS-GAIL}
  \label{algo:psgail}
  \begin{algorithmic}
    \STATE {\bfseries Input:} Expert trajectories $\tau_E \sim \pi_E$, Shared policy parameters $\Theta_0$, Discriminator parameters $\psi_0$, Trust region size $\Delta_{KL}$, Curriculum distribution $\mathcal{C}$
    \FOR{$k \gets 0, 1, \dotsc$}
    \STATE Sample number of agents from curriculum $m \sim \mathcal{C}(k)$
    \STATE Rollout trajectories for all $m$ agents $\vec{\tau} \sim \pi_{\theta_k}$
    \STATE Score $\vec{\tau}$ with critic, generating reward $\tilde{r}(s_{t},a_{t};\psi_k)$
    \STATE Batch trajectories obtained from all $m$ agents
    \STATE Take a TRPO step to find $\pi_{\theta_{k+1}}$ maximizing~\cref{eq:trpoobj}
    \STATE Update the critic parameters $\psi$ by maximizing~\cref{eqn:gail} 
    \ENDFOR
  \end{algorithmic}
\end{algorithm}

\section{Experiments}
\label{section:case_study}

\subsection{Simulator}
\label{subsection:simulator}
In order to learn the policy in an environment with human drivers, we use a simulator that allows for playing back real trajectories and simulating the movement of controlled vehicles given actions selected by a policy. The process proceeds as follows:

\begin{enumerate}
\item The initial scene is sampled from a dataset of real driver trajectories. This state includes the position, orientation, and velocity of all vehicles in the scene.

\item A subset of the vehicles in the scene are randomly selected to be controlled by the policy. For single-agent training only one vehicle is selected, whereas for multi-agent training $M$ vehicles are controlled by the policy.

\item For each vehicle, a set of features are extracted and passed to the policy as the observation. Table \ref{table:feats} describes the features provided to the policy. These features represent the scene information, and thus act as observations of the state of the driving MDP.
  A set of lidar-like beams emanating from the vehicle are used to gather information about its surroundings. We have 20 beams, each of which gives us the range and range rate of the first vehicle struck by them.

\item At every timestep, the policy outputs longitudinal acceleration and turn-rate values as the vehicle action in response to the observed features. These values are used to propagate the vehicle forward in time according to the vehicle dynamics.

\item The simulation is carried out, and associated metrics of both imitation performance and driving performance are extracted.
\end{enumerate}

\begin{table}[t]
\centering
\caption{Observation features}
\label{table:feats}
\begin{tabular}{@{}ll@{}}
\toprule

\textbf{Feature} & \textbf{Description} \\ \midrule

    Ego Vehicle 
    & Lane-relative velocity, heading, offset. \\ 
    & Vehicle length and width. \\ 
    & Longitudinal and lateral acceleration. \\
    & Local and global turn and angular rate. \\ 
    & Lane curvature, distance to left and \\
    & right lane makers and road edges. \\ \midrule

    Surrounding Vehicles
	& \num{20} artificial LIDAR beams \\
    & output in regular polar intervals, \\ 
    & providing the relative position  \\ 
    & and velocity of intercepted objects. \\ \midrule

	Temporal & Timegap and time-to-collision. \\ \midrule

	Indicators  & Collision occurring, ego vehicle \\
    & out-of-lane, and negative velocity. \\ 

\bottomrule
\end{tabular}
\end{table}

\subsection{Policy Representation}
Our learned policy must be able to simulate human driving behavior, which involves:
\begin{itemize}
  \item \textit{Non-linearity} in the desired mapping from states to actions (e.g., large corrections in steering to avoid collisions caused by small changes in the current state).
  \item \textit{High-dimensionality} of the state representation, which must describe properties of the ego-vehicle, in addition to surrounding cars and road conditions.
  \item \textit{Stochasticity} because humans may take different actions each time they encounter a given traffic scene.
\end{itemize}

To address the first and second points, we represent all learned policies $\pi_{\theta}$ using neural networks. Neural networks have gained widespread popularity due to their ability to learn robust hierarchical features from complicated inputs~\citep{lee2009convolutional,krizhevsky2012imagenet}, and have been used in automotive behavioral modeling for action prediction in car-following contexts~\citep{hongfei2003develop,panwai2007neural,khodayari2012modified,lefevre2014comparison,morton2016analysis}, lateral position prediction~\citep{liu2014latpos}, and maneuver classification~\citep{boyraz2007signal}.

To addresss the third point, we interpret the network's real-valued outputs given input $s_t$ as the mean $\mu_t$ and logarithm of the diagonal covariance $\log \nu_t$ of a Gaussian distribution. 
This enables stochasticity in the driving action provided by the neural network policy in response to a particular driving scene. Actions are chosen by sampling $a_t \sim \pi_{\theta}(a_t \mid s_t)$.

We evaluate both feedforward and recurrent network architectures. Feedforward neural networks directly map inputs to outputs.
The most common architecture, multilayer perceptrons (MLPs), consist of alternating layers of tunable weights and element-wise nonlinearities. However, the feedforward MLP is limited in its ability to adequately address partially observable environments.
In real world driving, sensor error and occlusions may prevent the driver from seeing all relevant parts of the driving state.
By maintaining sufficient statistics of past observations in memory, recurrent policies~\citep{wierstra2010recurrent} disambiguate perceptually similar states by acting with respect to histories of, rather than individual, observations.
We represent recurrent policies using Gated Recurrent Unit (GRU) networks~\citep{cho2014learning} due to their comparable performance with fewer parameters than other architectures.

We use similar architectures for the feedforward and recurrent policies. 
The recurrent policies consist of five feedforward layers that decrease in size from 256 to 32 neurons, with an additional GRU layer consisting of 32 neurons. 
Exponential linear units (ELU) were used throughout the network, which have been shown to combat the vanishing gradient problem while supporting a zero-centered distribution of activation vectors.
The MLP policies have the same architecture, except the GRU layer is replaced with an additional feedforward layer.  For each network architecture, one policy is trained through BC and one policy is trained through GAIL. In all, we trained four neural network policies: GAIL GRU, GAIL MLP, BC GRU, and BC MLP.

\subsection{Metrics}
\label{subsection:metrics}
~\Cref{fig:imitation_pipeline} shows the imitation learning pipeline starting from driving demonstration data to driving policies. 
We assess the imitation performance of our driving policies via different metrics. 
First, to measure imitation of local vehicle behaviors, we use a set of Root Mean Square Error (RMSE) metrics that quantify the distance between the real trajectories in the dataset and the trajectories generated by our learned driving policies. 
We calculate the RMSE between the original human driven vehicle and its replacement policy driven vehicle in terms of the position, speed, and lane offset.

Second, to assess the undesirable traffic phenomena that arise out of vehicular interactions as compared to local, single vehicle imitation, we extract metrics that quantify collisions, hard-braking, and offroad driving.
We also extract these metrics of undesirable traffic phenomena for the NGSIM driving data and compare them against the metrics obtained from rollouts generated by our driving policies.
\subsection{Single Agent Imitation}
\label{single_agent}
First, we report results obtained from experiments conducted on learning driving from a single agent~\citep{kuefler2017imitating}. Here, one vehicle is randomly sampled from the NGSIM demonstration data and its trajectory is used to train the critic. The effectiveness of the resulting driving policy trained using GAIL in imitating human driving behavior is assessed by validation in rollouts conducted on the simulator described in \cref{subsection:simulator}. 
The resulting driving behavior was compared against various driver modeling baselines using the metrics discussed in \cref{subsection:metrics}.

The first baseline is a static Gaussian (SG) model, which is an unchanging Gaussian distribution $\pi(a \mid s) = \mathcal{N}(a \mid \mu, \Sigma)$ fit using maximum likelihood estimation on the demonstration data.
The second baseline model is a Behavioral Cloning (BC) approach using mixture regression (MR)~\citep{lefevre2014comparison}.
The model has been used for model-predictive control and has been shown to work well in simulation and in real-world drive tests.
Our MR model is a Gaussian mixture over the joint space of the actions and features, trained using Expectation Maximization.
The stochastic policy is formed from the weighted combination of the Gaussian components conditioned on the features.
Greedy feature selection is used during training to select a subset of predictors up to a maximum feature count threshold while minimizing the Bayesian information criterion~\citep{schwarz1978estimating}.

The final baseline model uses a rule-based controller to govern the lateral and longitudinal motion of the ego vehicle.
The longitudinal motion is controlled by the Intelligent Driver Model~\citep{treiber2000congested}.
The inputs to the model are the vehicle's current speed $v(t)$ at time $t$, relative speed $r(t)$ with respect to the leading vehicle, and distance headway $d(t)$.
The model then outputs an acceleration according to
\begin{equation}
   a_{\mathrm{IDM}} = a_{\mathrm{max}}\Bigg( 1-\bigg(\frac{v(t)}{v_{\mathrm{des}}} \bigg)^4 - \bigg( \frac{d_{\mathrm{des}}}{d(t)} \bigg)^2 \Bigg) \text{,}
   \label{eqn:a_idm}
\end{equation}
where the desired distance is
\begin{equation}
    d_{\mathrm{des}} = d_{\mathrm{min}} + \tau .v(t) - \frac{v(t)\cdot r(t)}{2\sqrt{a_{\mathrm{max}}\cdot b_{\mathrm{pref}}}} \text{.}
    \label{eqn:d_des}
\end{equation}

The model has several parameters that determine the acceleration output based on the scene information. Here, $v_{\mathrm{des}}$ refers to the free speed velocity, $d_{\mathrm{min}}$ refers to the minimum allowable separation between the ego and leader vehicle, $\tau$ refers to the minimum time separation allowable between ego and leader vehicle, $a_{\mathrm{max}}$ and $b_{\mathrm{pref}}$ refer to the limits on the acceleration and deceleration, respectively.
For lateral motion, MOBIL~\citep{kesting2007general} is used to select the desired lane, with a proportional controller used to track the lane centerline.
We add noise to both the lateral and longitudinal accelerations to make the controller nondeterministic.

\begin{figure}
\begin{center}
    \includegraphics[width=\columnwidth]{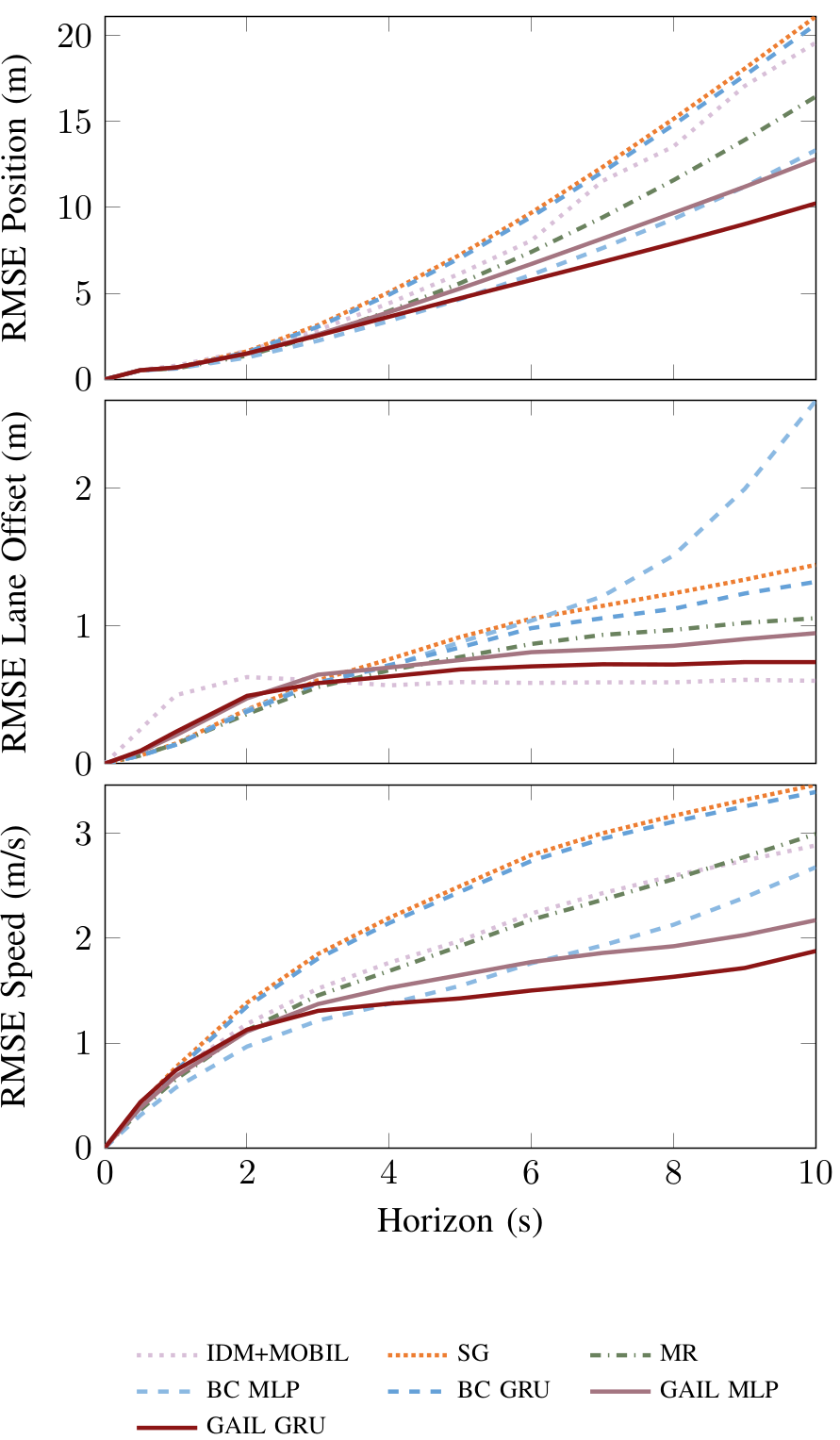}
\end{center}

  \caption{The root mean square error in position, velocity and lane offset for each candidate model versus prediction horizon. Policies trained using GAIL outperform the other methods.
  }
  \label{fig:rmse_single}
\end{figure}

To extract the metrics of driving performance, the ego vehicle is driven using a bicycle model with acceleration and turn rate sampled from the policy network.
\Cref{fig:rmse_single} shows the discrepancy between rollouts and the ground truth demonstration through root mean square error metrics.
The RMSE results show that the BC models have competitive short-horizon performance, but accumulate error over longer time horizons.
GAIL produces more stable trajectories and its short term predictions perform well.

\begin{figure}
\begin{center}
    \includegraphics[width=\columnwidth]{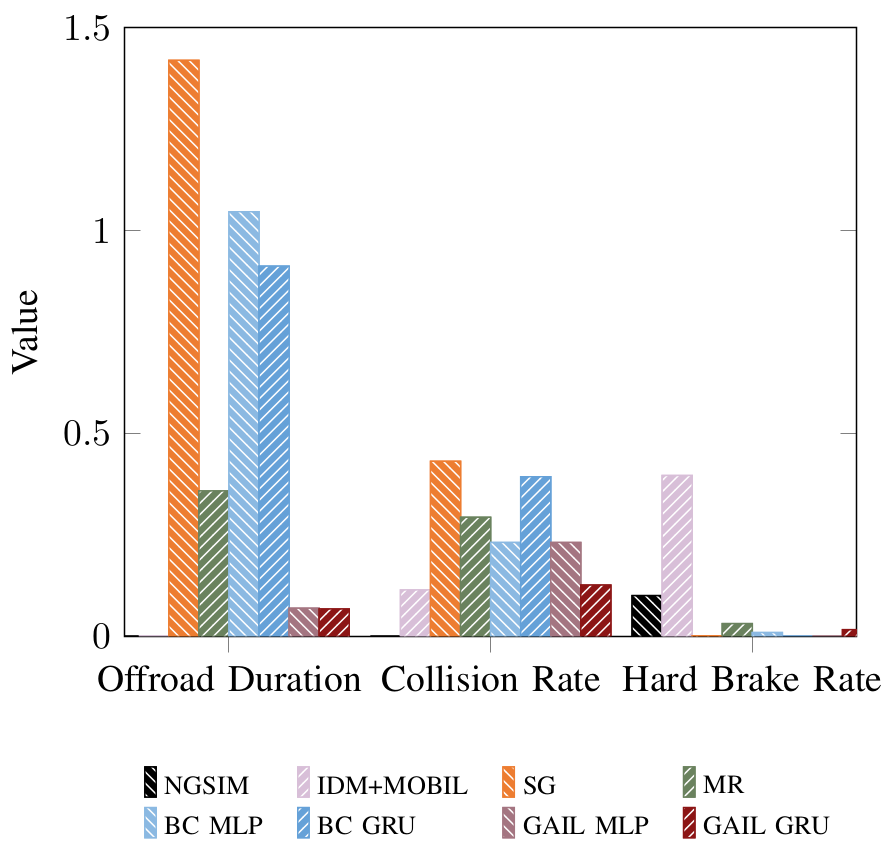}
\end{center}
  
  \caption{Metrics of undesirable traffic phenomena such as collisions, off the road driving, and hard decelerations. The GAIL based driving policies perform better than BC.
  }
  \label{fig:undesirable_single}
\end{figure}

\Cref{fig:undesirable_single} shows the undesirable driving metrics obtained from simulation. The GAIL policies outperform the BC policies.
Compared to BC, the GAIL GRU policy has the closest match to the data everywhere except for hard brakes, as it rarely takes extreme actions.
Mixture regression largely performs better than SG and is on par with the BC policies, but is still susceptible to cascading errors.
Offroad duration is perhaps the most striking statistic; only GAIL (and of course IDM + MOBIL) stay on the road for extended stretches.
SG never brakes hard as it only drives straight, causing many collisions as a consequence.
It is interesting that the collision rate for IDM + MOBIL is roughly the same as the collision rate for GAIL GRU, despite the fact that IDM + MOBIL should not collide. The inability of other vehicles within the simulation environment to fully react to the ego-vehicle may explain this phenomenon.

The results demonstrate that GAIL-based models capture many desirable properties of both rule-based and machine learning methods, while avoiding common pitfalls.
With the exception of the rule-based controller, GAIL policies achieve the lowest collision and off-road driving rates, considerably outperforming baseline and similarly structured BC models. Furthermore, extending GAIL to recurrent policies leads to improved performance.
This result is an interesting contrast with the BC policies, where the addition of recurrence tends not to yield better results.
Thus, we find that recurrence by itself is insufficient for addressing the detrimental effects that cascading errors can have on BC policies.
\subsection{Multi Agent Imitation}
\label{multi_agent}
%\Cref{single_agent} described the driving behavior of a single vehicle in traffic scenes populated by demonstrated driving trajectories from the NGSIM dataset. 
In this subsection, we describe experiments and results conducted for multiple learning agents using the parameter sharing approach (PS-GAIL) described in~\Cref{section:multi-agent}.

In the multi-agent setting, multiple vehicles are sampled from the demonstration NGSIM data, and a policy with shared parameters is learned by batching together the observations and actions from all the vehicles. Importantly, the dynamics of the environment change along with the agent policies. Our training procedure must therefore account for non-stationary environment dynamics.

We mitigate this problem by introducing a curriculum that scales the difficulty of the multi-agent learning problem during training. \citet{gupta2017cooperative} define a multi-agent curriculum, $\mathcal{C}$, as a multinomial distribution over the number of agents controlled by the policy each episode. The curriculum gradually shifts probability mass to larger numbers of agents. We use a simplified curriculum that increments the number of controlled agents by a fixed number every $K$ iterations during training, in which case $\mathcal{C}(k)$ is a deterministic function of the iteration $k$.

We use recurrent neural network (RNN) policies, in all cases consisting of \num{64} Gated Recurrent Units (GRUs). The observation is passed directly into the RNN without any initial reduction in dimensionality. We use recurrent policies in order to address the partial observability of the state caused by occluded vehicles. In the multi-agent setting, a single shared policy selects actions for all vehicles, following the parameter sharing approach previously described. Policy optimization is performed using an implementation of TRPO from rllab~\citep{duan2016benchmarking} with a step size of \num{0.1}.

We use two training phases. The first phase consists of \num{1000} iterations with a low discount of \num{0.95} and a small batch size of \num{10000} observation-action pairs. The second phase fine-tunes the models, running for \num{200} iterations with a higher discount of \num{0.99} and larger batch size of \num{40000}.
For the multi-agent model, we add \num{10} agents to the environment every \num{200} iterations of the first training phase. 
We use \num{100} agents in the fine-tune phase for the multi-agent GAIL models.

The critic acts as the surrogate reward function in the environment. The observation-action pairs for each vehicle at each timestep are passed to the critic, which outputs a scalar value that is then used as the reward for that vehicle. The critic is implemented as a feed-forward neural network consisting of (\num{128},\num{128},\num{64}) ReLU units. We implemented the critic as a Wasserstein GAN with a gradient penalty (WGAN-GP) of \num{2}~\citep{gulrajani2017improved}. 
Similar to ~\citet{li2017infogail}, we used a replay memory for the critic in order to stabilize training, which contains samples from the three most recent epochs. 
For each training epoch of the policy, the critic is trained for \num{40} epochs using the Adam optimizer~\citep{kingma2014adam} with a learning rate of \num{0.0004}, dropout probability of \num{0.2}, and batch size of \num{2000}. 
Half of each batch consists of NGSIM data, with the remaining half comprised of data from policy rollouts. 
Finally, the reward values output from the critic are adaptively normalized to have zero mean and unit variance prior to being passed to TRPO.

\begin{figure}[h]
\begin{center}
    \includegraphics[width=\columnwidth]{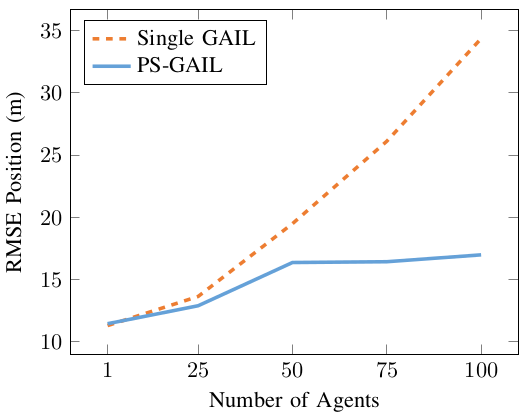}
\end{center}
    \caption{Average RMSE position value across all timesteps of an episode as a function of the number of controlled agents. As the policy controls more vehicles, single-agent GAIL performance deteriorates rapidly, while PS-GAIL performance decays more slowly.
    }
    \label{fig:scaling}
\end{figure}

The difficulty of the multi-agent task scales with the number of agents controlled in the environment. 
\Cref{fig:scaling} shows the performance of the two models as a function of the number of agents driven by our learned driving policy. 
The indicated number of agents are randomly sampled and replaced in the environment with the policy, while the remaining agents are left as originally recorded in NGSIM. 
Here, the single-agent policy refers to the policy trained using data obtained from one vehicle, and then deployed on multiple vehicles during validation. 
The results indicate that while the single-agent policy deteriorates rapidly with increasing number of agents, the multi-agent policy declines in performance gradually.

\subsubsection{Reward Augmentation}
Both single-agent GAIL and PS-GAIL are methodologies that are domain agnostic. However, for the specific task of driver modeling, providing the learning agent with domain knowledge proves useful. 
Reward Augmented Imitation Learning (RAIL) provides external penalties during training~\citep{bhattacharyya2019simulating} that specifically encapsulate rules of the road. 
These include penalties for going off the road, braking hard, and colliding with other vehicles. 
All of these are undesirable driving behaviors and therefore should be discouraged in the learning agent. 
These penalties help to improve the state space exploration of the learning agent by discouraging bad states such as those that could potentially lead to collisions. 
In RAIL, part of the reinforcement learning cost signal comes from the critic based on imitating the expert, and another cost signal comes from the externally provided penalties specifying the prior knowledge of the expert~\citep{li2017infogail}.

We explore a binary penalty and a smoothed penalty as the two forms of reward augmentation provided to the imitation learning agent.
The first method of reward augmentation that we employ is to penalize states in a binary manner, where the penalty is applied when a particular event is triggered.
To calculate the augmented reward, we take the maximum of the individual penalty values.
For example, if a vehicle is driving off the road and colliding with another vehicle, we only penalize the collision. 
This will also be important when we discuss \textit{smoothed penalties}.

We explore penalizing three different behaviors. 
First, we give a large penalty $R$ to each vehicle involved in a collision.
Next, we impose the same large penalty $R$ for a vehicle that drives off the road. 
Finally, performing a hard brake (acceleration of less than \SI{-3}{\meter\per\second^{2}}) is penalized by only ${R}/{2}$.
The penalty formula is shown in~\cref{binaryreward}.
We denote the smallest distance from the ego vehicle to any other vehicle on the road as $d_c$ (meters), where $d_c \geq 0$.
% If $d_c < 0$, then we know two cars are overlapping.
We also define the closest distance from the ego vehicle to the edge of the road (meters): $d_{\text{road}} = \min\{d_{\text{left}}, d_{\text{right}}\}$. 
We allow $d_{\text{road}}$ to be negative if the vehicle is off the road. 
Finally, let $a$ be the acceleration of the vehicle in \SI{}{\meter\per\second^2}. A negative value of $a$ indicates that the vehicle is braking.
Now, we can formally define the binary penalty function:
\begin{align}
\label{binaryreward}
    \text{Penalty} & = 
    \begin{cases}
       R & \mathrm{if} \; d_c = 0 \\
       R & \mathrm{if} \; d_{\text{road}} \leq -0.1 \\
       \frac{R}{2} & \mathrm{if} \; a \leq -3 \\
       0 & \; \mathrm{otherwise}
    \end{cases} 
\end{align}
% Both collisions and off-road driving receive a penalty of $R$. 
The relative values of the penalties indicate the preferences of the designer of the imitation learning agent.
For example, in this case study, we penalize hard braking less than the other undesirable traffic phenomena.

We hypothesize that providing advanced warning to the imitation learning agent in the form of smaller, increasing penalties as the agent approaches an event threshold will address the credit assignment problem in reinforcement learning. %cite
We provide a \textit{smooth penalty} for off-road driving and hard braking, where the penalty is linearly increased from a minimum threshold to the previously defined event threshold for the binary penalty. 
For off-road driving, we linearly increase the penalty from $0$ to $R$ when the vehicle is within \SI{0.5}{\meter} of the edge of the road.
For hard braking, we linearly increase the penalty from $0$ to $R/2$ when the acceleration is between \SI{-2}{\meter\per\second^{2}} and \SI{-3}{\meter\per\second^{2}}.

\begin{figure}
\begin{center}
    \includegraphics[width=\columnwidth]{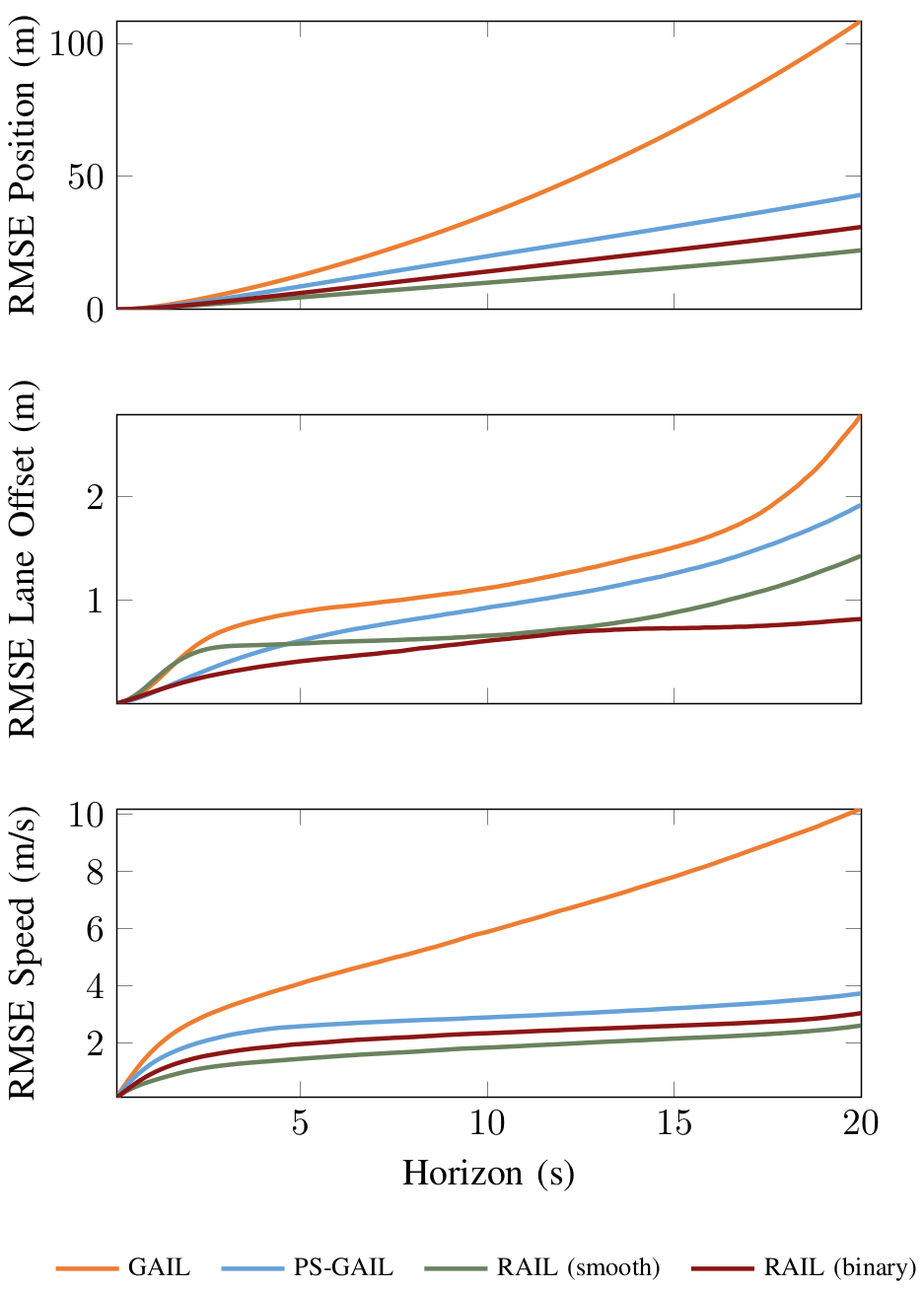}
\end{center}
\caption{A comparison of root mean square error in position, lane offset and speed for single-agent, multi-agent and reward augmented GAIL versus prediction horizon. Policies trained using reward augmented GAIL show better performance.}
\label{fig:rmse_psgail_rail}
\end{figure}

The driving performance of driving policies trained using PS-GAIL and RAIL was assessed by performing experiments in the simulator. 
\Cref{fig:rmse_psgail_rail} shows root mean square error results for prediction horizons up to \SI{20}{\second}.
These plots indicate that the multi-agent learning approaches PS-GAIL and RAIL capture expert behavior more faithfully than single-agent GAIL.
This performance discrepancy is especially pronounced for longer prediction horizons, where the errors for single-agent policies begin to accumulate rapidly.
Further, reward augmentation results in better local imitation performance, as seen by the lowest RMSE values.

\begin{figure}
\begin{center}
    \includegraphics[width=\columnwidth]{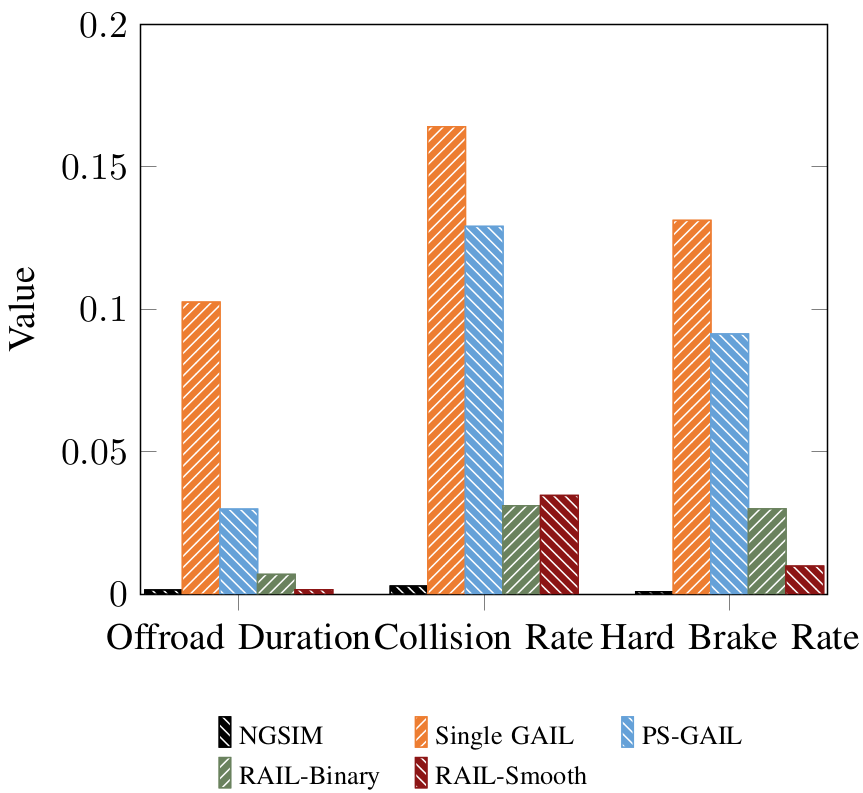}
\end{center}
  \caption{Metrics of undesirable traffic phenomena. These are explicitly penalized in the reward augmentation formulation. RAIL results in policies with lower collisions, offroad driving and hard braking as compared to the PS-GAIL baseline.
  }
  \label{fig:undesirable_psgail_rail}
\end{figure}

The superior performance of PS-GAIL and RAIL is further illustrated by~\Cref{fig:undesirable_psgail_rail}.
These validation results empirically demonstrate that PS-GAIL and RAIL policies are less likely to lead vehicles into collisions, extreme decelerations, and off-road driving.
This shows that the PS-GAIL training procedure encourages stabler interactions between agents, thereby making them less likely to encounter extreme or unlikely driving situations. 
The inclusion of domain knowledge is especially significant here as seen by the reduction in the values of the undesirable metrics of driving.

\subsection{Disentangling Driving Styles}
Human driving demonstrations display variability due to latent factors. In this subsection, we report results from experiments targeted at disentangling driving styles from demonstrations~\citep{kuefler2018burn}.

\begin{figure}
\begin{center}
\includegraphics[scale=0.2]{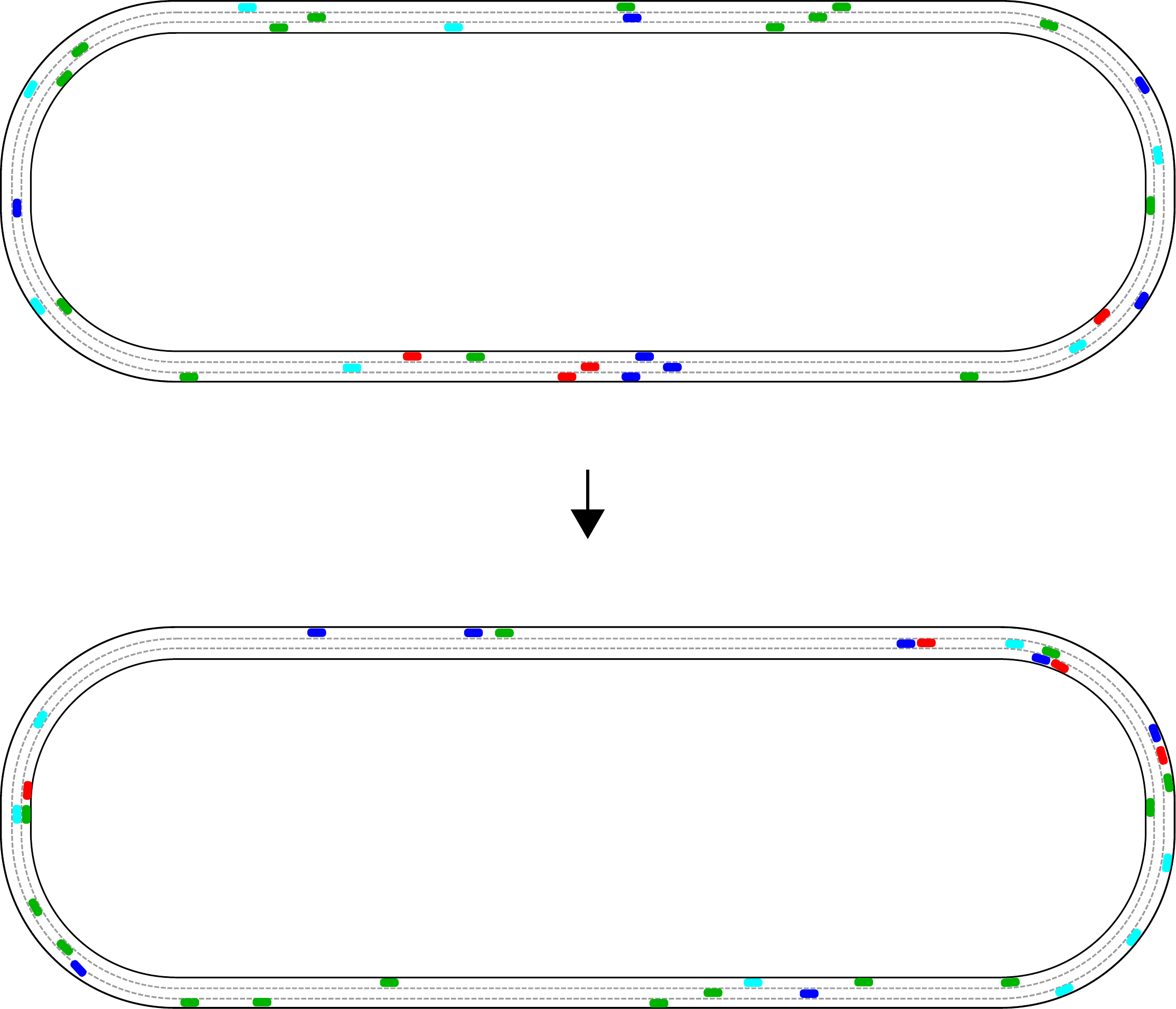}
\caption{Scenes taken from oval track environment after initialization and a few seconds of driving. Over time, tail-gaiters (green) and aggressive drivers (red) cluster behind passive drivers (blue). Speeders (cyan) retain their large headway distances.}
\label{fig:tracks}
\end{center}
\end{figure}

The simulator used to generate data and train models is based on an oval racetrack, shown in ~\cref{fig:tracks}~\citep{morton2017simultaneous}. We populate our environment with vehicles simulated by the Intelligent Driver Model \citep{treiber2000congested}, where lane changes are executed by the MOBIL general lane changing model \citep{kesting2007general}. The settings of each controller are drawn from one of four possible parameterizations, defining the style $z$ of each car. The resulting driving experts fall into one of four classes:

\begin{itemize}

\item Aggressive: High speed, large acceleration, small headway distances.

\item Passive: Low speed, low acceleration, large headway distances.

\item Speeder: High speed and acceleration, but large headway distance.

\item Tailgating: Low speed and acceleration, but small headway distances.
\end{itemize}

Furthermore, the desired speed of each car is sampled from a Gaussian distribution, ensuring that individual cars belonging to the same class behave differently. A total of 960  training demonstrations and 480 validation demonstrations were used, each lasting 50 timesteps (or 5 seconds, at 10 Hz). The observations are represented with the combination of LIDAR and road features reported in \cref{table:feats}.

We compared against three baseline models. The first baseline is the VAE driver policy proposed by \citeauthor{morton2017simultaneous} \citeyear{morton2017simultaneous}. Its encoder network consists of two Long Short-Term Memory (LSTM) \citep{hochreiter1997long} layers that map state-action pairs to the mean and standard deviation of a 2-dimensional Gaussian distribution. Its decoder, or policy, is a 2-layer MLP, also consisting of 128 units. During testing, the encoder conditions on a sequence of observation-action pairs sampled from the expert whose playback is used to initialize the ego vehicle (the ``burn-in demonstration''). The predicted mean of the distribution is used as the latent code for the policy. The second baseline is a GAIL model trained on the demonstration trajectories. It has the same model architecture as $\pi_\theta$, the policy trained using Burn-InfoGAIL with the exclusion of the learned embedding layer needed to encode the latent style variable. Finally, we baseline against an implementation of InfoGAIL that is architecturally identical to $\pi_\theta$, but simply samples $z$ from a discrete uniform distribution at the beginning of each trial.

% \begin{figure}[h]
% \begin{center}
% \includegraphics[height=0.8\columnwidth,width=0.7\columnwidth]{figs/rmse}
% \caption{Root mean squared error (RMSE) between learned policies and validation trajectories. Results are averaged over \num{1000} rollouts for each model. Our model achieves the lowest error on predicting both speed and position over 30 second trajectories.}
% \label{fig:rmse}
% \end{center}
% \end{figure}
\begin{figure}
\begin{center}
    \includegraphics[width=\columnwidth]{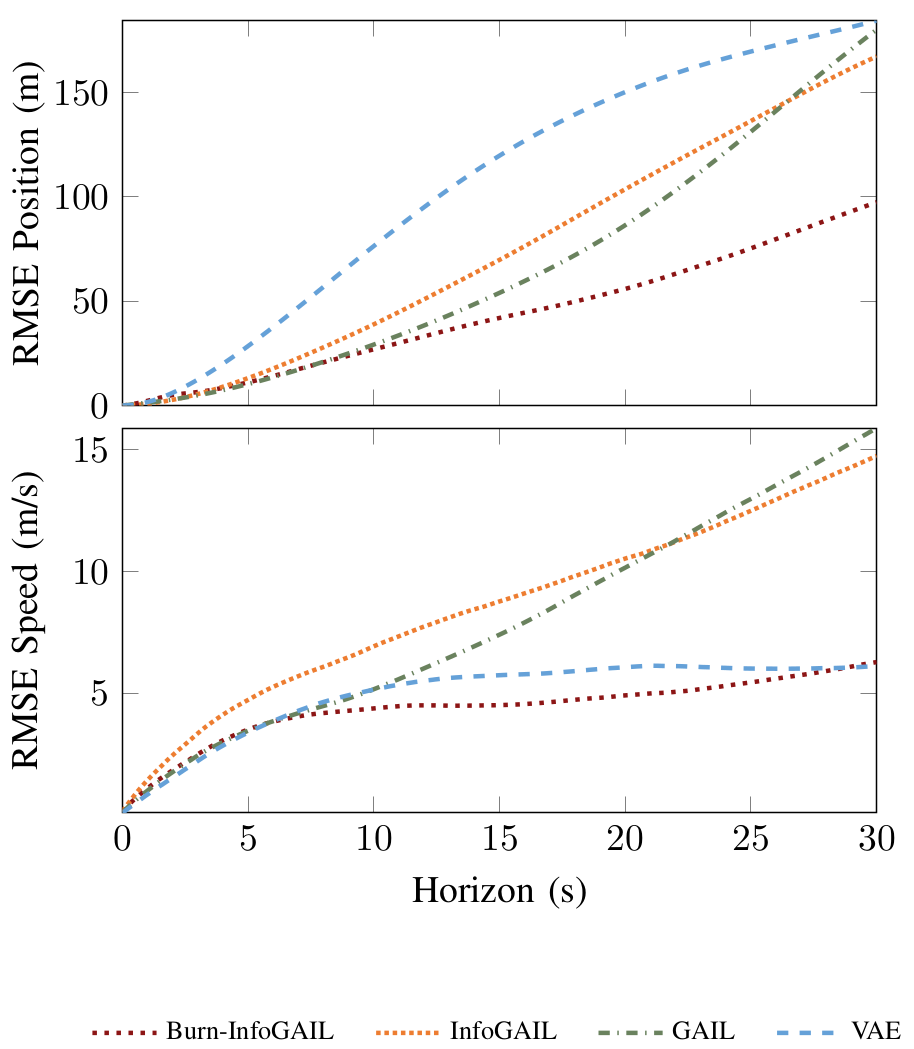}
\end{center}
  \caption{Root mean squared error (RMSE) between learned policies and validation trajectories. Results are averaged over \num{1000} rollouts for each model. Our model achieves the lowest error on predicting both speed and position.
  }
  \label{fig:rmse_burninfogail}
\end{figure}

As shown in \Cref{fig:rmse_burninfogail}, Burn-InfoGAIL achieves the lowest error over the longest period of driving. 
GAIL is able to capture differences in style for about \SI{10}{\second}, presumably because the imitation objective discourages the policy from adjusting its velocity away from its initial conditions. 
As minor errors compound over long horizons, GAIL drifts toward an average policy due to its mode-seeking nature~\citep{goodfellow2014generative}. 
In contrast, the VAE is able to use the latent code inferred from the burn-in demonstration to maintain an appropriate speed, achieving an RMSE close to the true value, rivaling Burn-InfoGAIL. 
However, being trained without a simulator, the VAE suffers from cascading errors causing it to go off the road.

\section{Conclusion}
\label{section:conclusion}
Reliable models of human driving are essential for the safety validation of autonomous driving algorithms.
    In this paper, we model human driving as a sequential decision making problem under uncertainty.
        The problem has continuous state and action spaces, non-linearity, stochasticity and an unknown cost function.

Learning from demonstrations is a promising approach to solving MDPs when the cost function is unknown or difficult to specify. Following on a long line of work on inverse reinforcement learning, GAIL was proposed with the promise of (in theory) exact imitation even for problems with large (potentially continuous) state and actions spaces. Driver modeling is a problem where the state and action spaces are continuous, the policy is characterized by non-linearity and stochasticity, and the cost function is difficult to articulate exactly. These characteristics make learning from human driving demonstrations a suitable approach to generating human-like driving behavior. However, human demonstrations are dependent on latent factors of variability that cannot be captured by GAIL on its own. Moreover, driver modeling is a multi-agent problem not directly solvable by GAIL. 

This article described three modifications to GAIL addressing these limitations. First, it described PS-GAIL, which accounts for the multi-agent nature of the problem resulting from the interaction between traffic participants. Second, it described RAIL, which uses reward augmentation to provide domain knowledge about the rules of the road to the driver modeling agent. Third, it described Burn-InfoGAIL which deals with the disentanglement of latent variability in demonstrations. All three modifications were demonstrated on driver modeling experiments, including learning driver behavior models from real world driving demonstration data.

Directions for future work include methods for improving model performance, and applying learned driver models. Potential methods for improving model performance include (i) explicitly modeling the interaction between agents in a centralized manner through the use of Graph Neural Networks \citep{battaglia2018relational} and coordination graphs~\citep{kuyer2008multiagent,li2021deep}, (ii) exploring recently introduced, alternative methods of imitation learning \citep{reddy2019sqil}, and (iii) scaling up experiments to larger datasets and driving domains.
Ultimately, the goal in learning human driver models is to validate autonomous vehicles in simulation, and we hope to apply these models to that end in the future.

% \section*{Acknowledgments}
% Toyota Research Institute (TRI) provided funds to assist the authors with their research, but this article solely reflects the opinions and conclusions of its authors and not TRI or any other Toyota entity.

% \section*{REFERENCES}
% \renewcommand{\section}[2]{}%
% \bibliography{ref.bib}

% \clearpage
% \newpage

% \bibliographystyle{IEEEtran}
% \bibliography{ref}

\renewcommand*{\bibfont}{\footnotesize}
\printbibliography

% \addtolength{\textheight}{-12cm}   
% This command serves to balance the column lengths on the last page of the document manually. It shortens the textheight of the last page by a suitable amount. This command does not take effect until the next page so it should come on the page before the last. Make sure that you do not shorten the textheight too much.

% \vspace{1cm}
% \newpage

\end{document}